\documentclass[letterpaper, 10 pt, conference]{ieeeconf}  

\usepackage{graphicx}
\usepackage{float}

\IEEEoverridecommandlockouts                              

\title{\LARGE \bf
Industry 6.0: New Generation of Industry driven by Generative AI and Swarm of Heterogeneous Robots
}

\author{Artem Lykov$^{*}$, Miguel Altamirano Cabrera$^{*}$, Mikhail Konenkov$^{*}$, Valerii Serpiva, \\ Koffivi Fidèle Gbagbe, Ali Alabbas, Aleksey Fedoseev, Luis Moreno, \\ Muhammad Haris Khan, Ziang Guo, and Dzmitry Tsetserukou
\thanks{* Denotes equal contribution.}
\thanks{**This work was not supported by any organization.}
\thanks{Authors are with Intelligent Robotics Laboratory, CDE, Skoltech,
         Bolshoy Boulevard, 30, bld. 1, Moscow 121205, Russia
        {\tt\small \{Artem.Lykov, M.Altamirano, Mikhail.Konenkov, Valerii.Serpiva, Koffivi.Gbagbe, Ali.Alabbas, Aleksey.Fedoseev, Luis.Moreno, Haris.Khan, Ziang.Guo, D.Tsetserukou\}@skoltech.ru}}%
}

\begin{document}

\maketitle
\thispagestyle{empty}
\pagestyle{empty}



\begin{abstract}

This paper presents the concept of Industry 6.0, introducing the world's first fully automated production system that autonomously handles the entire product design and manufacturing process based on user-provided natural language descriptions. By leveraging generative AI, the system automates critical aspects of production, including product blueprint design, component manufacturing, logistics, and assembly. A heterogeneous swarm of robots, each equipped with individual AI through integration with Large Language Models (LLMs), orchestrates the production process. The robotic system includes manipulator arms, delivery drones, and 3D printers capable of generating assembly blueprints. The system was evaluated using commercial and open-source LLMs, functioning through APIs and local deployment. A user study demonstrated that the system reduces the average production time to 119.10 minutes, significantly outperforming a team of expert human developers, who averaged 528.64 minutes (an improvement factor of 4.4). Furthermore, in the product blueprinting stage, the system surpassed human CAD operators by an unprecedented factor of 47, completing the task in 0.5 minutes compared to 23.5 minutes. This breakthrough represents a major leap towards fully autonomous manufacturing. 

\end{abstract}


\section{Introduction} 


The rapid advancement of generative AI has significantly accelerated progress in cognitive robotics, enabling success across various form factors. Notably, robotic manipulators have set new benchmarks in reasoning and real-world interaction. Examples such as Google DeepMind's PaLM-E \cite{driess2023palm}, RT-2 \cite{brohan2023rt}, and RT-X \cite{Neill_2024} demonstrate advanced capabilities in advanced interaction with robotic arms. Quadruped robots, e.g., CognitiveDog \cite{lykov2024cognitivedog}, exhibit enhanced mobility and interaction through their decision-making ability. Additionally, Tesla Optimus, Agility Robotics \cite{agilityrobotics}, and OpenAI \cite{figureAI} are increasingly integrating generative AI to enhance the performance of humanoid robots in a wide range of human-centered environments. Mobile robots, exemplified by \cite{lykov2023llm-mars} and \cite{hu2024deploying}, also showcase significant advancements. Recently, the Bi-VLA \cite{fidele2024bi} introduced a Vision-Language-Action model that enables bimanual robotic manipulation. This system allows robots to interpret human instructions and execute complex tasks, such as grasping, cutting, and placing objects. Using a Large Language Model (LLM), the Bi-VLA system translates natural language commands into executable actions through predefined API functions. 

\begin{figure}[t!]
\centerline{\includegraphics[width=0.49\textwidth]{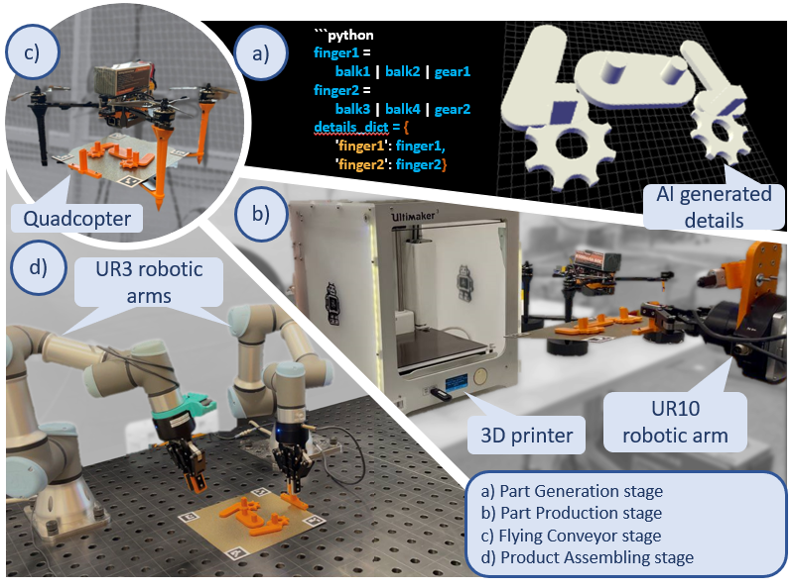}}
\caption{Implementation of Industry 6.0 concept.}
\label{fig:teaser}
\vspace{-0.4cm}
\end{figure}


In addition, LLMs have shown the capability to generate 3D objects from natural language descriptions provided by the user. It opens new directions for autonomous design and manufacturing. For example, Point-E \cite{nichol2022point} creates 3D point clouds from text using a diffusion model for shape refinement. G3PT systems \cite{zhang2024g3pt} utilize a cross-scale transformer to map point-based 3D data into hierarchical token sequences, improving generation quality. Additionally, Diffusion-SDF \cite{li2023diffusion} generates diverse 3D shapes with voxelized signed distance fields, enabling more detailed and flexible designs. The ability to generate 3D shapes with LLMs has been successfully applied to enhance the control of robot swarms, as demonstrated in FlockGPT \cite{lykov2024flockgpt}. This approach highlights the potential of LLMs to manage complex multi-agent environments. 

Additionally, incorporating LMMs represents a major leap forward in cognitive robotics, allowing robots to better understand and perform complex tasks across different systems. For example, the CognitiveOS  \cite{lykov2024cognitiveos} demonstrates the potential of LLMs to enhance robotic autonomy and adaptability, empowering robots with generative AI capabilities to manage manufacturing processes dynamically. Such advancements underscore the ongoing shift toward systems that perform predefined tasks and adapt and make informed decisions in real time.

As we continue to explore the capabilities of advanced AI, studies like those conducted by Fan et al. \cite{fan2024embodied} on leveraging LLMs for autonomous industrial robotics. They illustrate the substantial benefits these technologies bring to enhancing decision-making and task execution in robotics, thereby streamlining manufacturing processes. Similarly, ‘LLM-BRAIn’ \cite{lykov2023llm-brain} demonstrates how transformer-based LLMs can generate sophisticated robot behavior trees, aiding in autonomous robotic control and aligning with the evolving needs of fully automated manufacturing systems.

In the evolving field of autonomous manufacturing, using different types of robots together has become crucial for improving their ability to work effectively in constantly changing environments. With their sophisticated coordination among various robotic agents, these systems ensure that tasks are allocated and completed efficiently within complex manufacturing setups \cite{rizk2019cooperative, zlot2006market}. The rise of generative AI has propelled these capabilities even further, enabling real-time adjustments and solid decision-making crucial for contemporary manufacturing operations \cite{hagele2016industrial, heuss2023concept}. Notably, robots equipped with advanced perception technologies are better at using real-time data to improve accuracy and efficiency than classical automated machines, substantially cutting down on downtime and errors \cite{batra2021real, kappler2018real}. We're entering a new manufacturing era as AI technologies and autonomous robots continue to work more closely together. These processes require minimal human intervention and are optimized for precision and adaptability. 

These systems exhibit impressive performance individually, and ongoing research focuses on replacing certain human functions in manufacturing with robotic counterparts. Industry 5.0 mainly stands for intelligent human-robot collaborations leveraging safety provided by collaborative robots.  In this work, we suggest a new stage of the technology: Industry 6.0. In contrast to Industry 5.0, we suggest that the future industry will be fully driven by Generative AI and a swarm of heterogeneous robots without the involvement of human beings in the design and manufacturing process. Heterogeneous robots can include cobots, humanoid robots, mobile platforms, drones, and industrial robots. Each robot operates with a degree of autonomy, possessing its own intelligence, allowing it to adapt to changing conditions and make independent decisions. Our prototype accepts a user-defined description of a desired mechanism, generates a corresponding assembly, produces parts using 3D printers, transports components to an assembly station via an adaptive flying conveyor of drones, and assembles the product using robotic manipulators.


\section{System Architecture of Industry 6.0} 

\begin{figure}[htbp]
\vspace{+0.3cm}
\centerline{\includegraphics[width=0.47\textwidth]{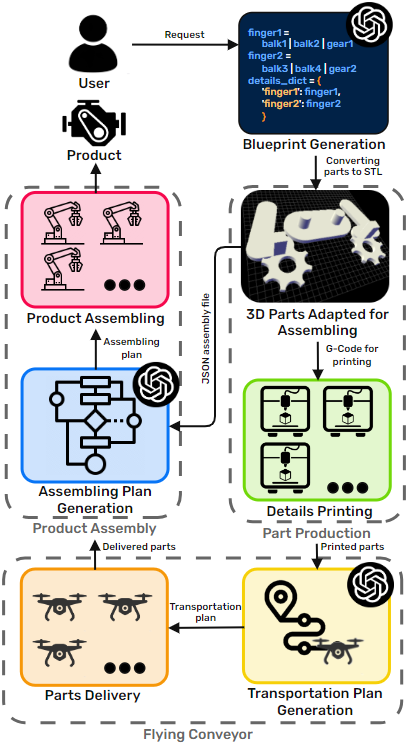}}
\caption{Industry 6.0 architecture, including stages of code generating, printing, delivering, and assembling required parts performed in the human-out-of-the-loop model.}
\label{fig:sys_arc}
\vspace{-0.5cm}
\end{figure}    

The Industry 6.0 system architecture (illustrated in Fig. \ref{fig:sys_arc}) describes how a user's natural language command is transformed into a physically manufactured product. The autonomous production system is composed of four primary components: assembly blueprint generation, part production, product assembly, and a flying conveyor system responsible for transporting components between these stages. In our project, the generative AI component is implemented via the OpenAI API \cite{openaiapi}. The local LLM inference option was also considered. LLM was integrated into the system using LangChain \cite{langchain}.

The AI accesses a list of available tools and resources used in the design process as context. The Retrieval-Augmented Generation (RAG) approach is employed for local model inference, while API-based models utilize all relevant information as contextual input. Based on the user’s request, the system constructs a 2D signed distance function (SDF), which determines the geometry of every part of the mechanism, defining the assembly geometry and specifying the necessary connection points. In the subsequent stage, the 2D SDF mechanism assembly is converted into a 3D STL assembly. Another STL file contains 3D parts of the mechanism spread out in the plane for 3D printing. The STL files are then further processed into G-code for 3D printing. Additionally, a JSON assembly file is generated, which stores information on parts' positioning in the assembly and on the print bed and details on all connection points and auxiliary elements for robotic grasping. While 3D printers are the primary fabrication tools in our system, other manufacturing cells, such as CNC machines, can also be integrated.


Printed components are delivered to the assembly station using drones. Using drones instead of traditional conveyor belts allows for adaptable production lines and the capability to transport parts in three-dimensional space, enabling the construction of vertical manufacturing chains. In a distributed system with multiple part production and assembly cells, the generative AI generates the transportation plan. In the present work, the task is simplified by having a single-part production cell and a single assembly cell.

Once all necessary components arrive at the assembly station, a step-by-step assembly plan is generated using generative AI based on the assembly file. This plan also accounts for the parameters of the assembly cell, such as the number of robotic manipulators available. The product is assembled following this plan, thus completing the autonomous production process.

\section{Product Manufacturing Driven by Generative AI}

\subsection{Assembly blueprint generation stage} \label{blueprint_generation_stage}

The blueprint generation for assembly marks is the initial and fundamental phase of the manufacturing process, laying the groundwork for the entire system. This stage takes a natural language description of the desired product as input. It outputs both a set of 3D-printable components (shown in Fig. \ref{fig:teaser}(a)) and a comprehensive file that specifies the initial and final positions of these components in the global coordinate system, along with the critical assembly elements within their respective local coordinate frames.

\begin{figure}[htbp]
\vspace{+0.2cm}
\centerline{\includegraphics[width=0.48\textwidth]{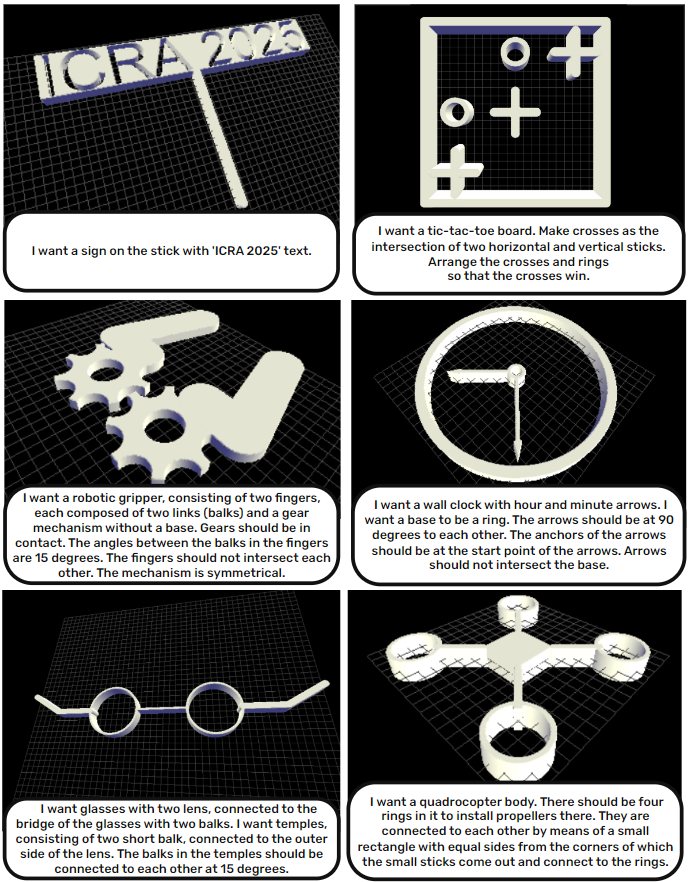}}
\caption{CAD model design by generative AI.}
\label{fig:sdf_examples}
\vspace{-0.2cm}
\end{figure}

The pipeline leverages LangChain — a framework for building controlled agentic workflows and splits the process into two tightly integrated sub-stages: user input analysis and code generation for mechanism SDF. Such decomposition ensures stable system behavior and improves the final result of the generation.

In the first sub-stage, the system enriches the user-provided input by providing a description of the mechanism assembly and operating principle. In this step, the LLM also describes the parts of the mechanism, the primitives that make up each part, and the connection between the parts necessary for the mechanism to operate. The list of 2D primitives is part of the LLM context and includes various 2D shapes, e.g., a balk, a circle, or a rectangle. The analysis sub-stage allows for the significant extension of information about the mechanism, which is critical for generating the code of the mechanism SDF.


In the second sub-stage, the LLM takes the refined product specification as input and develops the Python code to generate an SDF mechanism. Our SDF implementation is built upon an open-source Python library \cite{fogleman_sdf}. We developed a library of the mechanical primitives used to construct parts of commonly produced mechanisms, such as gears, beams, or rings. This expanded solution enables the creation of complex objects, geometric transformations, and 3D text. This set of primitives, along with an expertly curated set of rules, practices, and examples of the 2D mechanisms in Python code, is passed to the LLM context. The improved product description from the analysis sub-stage and excessive knowledge of SDF implementation allow the LLM to produce an accurate mechanism representation in executable Python code.

In the last stage of blueprint generation, the SDF mechanism representation produced by LLM is processed with the 3D STL generation script. In its current implementation, our prototype targets assemblies consisting of one or more components with limited mobility in two-dimensional space. Examples of product details in STL format generated in this step are shown in Fig. \ref{fig:sdf_examples}. To facilitate the assembly process, a base with pins is automatically generated. Upon generation of the components and the base in SDF format, the system automatically arranges them on the printing platform, adding functional pins for easy handling and assembly. Currently, the design supports single-layer assembly with a single type of dynamic connection—specifically, a pin-and-hole mechanism between the base and the components. Future iterations of this framework will introduce additional connection types and support for multi-layered assemblies. The final 3D models are exported in STL format, while a complementary JSON file is created to encapsulate the spatial configuration of the components and detailed descriptions of their fastening and functional elements.

\begin{figure}[t]
\vspace{+0.3cm}
\centerline{\includegraphics[width=0.48\textwidth]{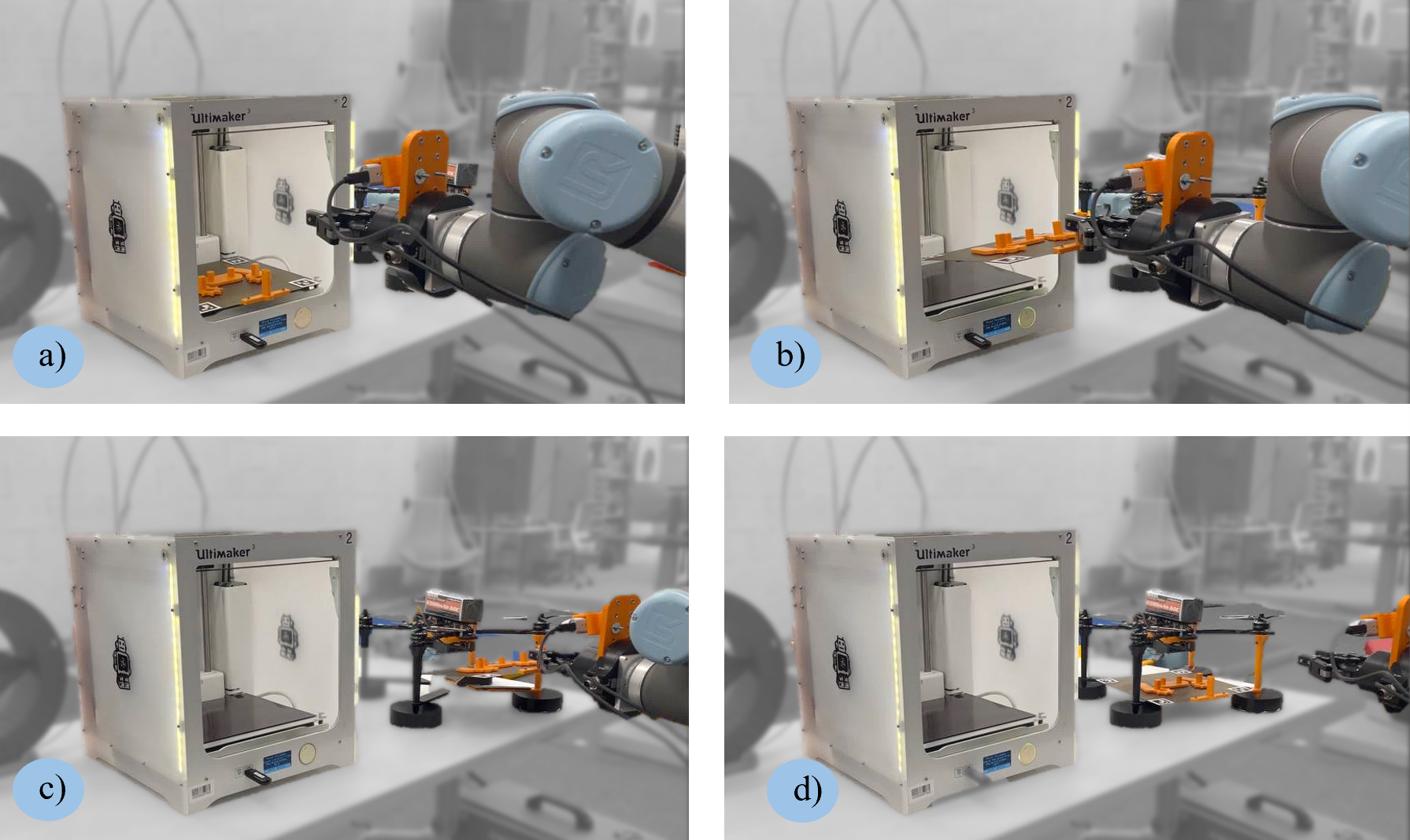}}
\caption{Transfer of the 3D printing surface from the printer to the drone. In (a), the Universal Robot UR10 grasps the printing surface, takes it out from the 3D printer (b), transfers (c), and places it on the drone (d).}
\label{fig:tranfer_printer_drone}
\vspace{-0.2cm}
\end{figure}






\subsection{Part production stage} 

Upon completion of the STL file generation is completed, the 3D models are converted into G-code with specific printing parameters for the Ultimaker 3D printer and subsequently sent for production. During the G-code generation, we deliberately exclude support structures and raft layers, as post-processing will occur without human intervention. This approach eliminates the need for manual removal of extraneous elements, optimizing the automated workflow.

The setup leverages a specialized magnetic printing surface equipped with a custom holder. This surface can be inserted and removed from the printer autonomously using a manipulator, ensuring secure attachment without bolts or latches. The surface is pre-aligned for precise printing, with AruCo markers placed at its corners. While the printer operates independently of these markers, they are critical post-print for identifying the coordinate system in which the printing was executed. This allows for accurate spatial localization of the printed parts in subsequent stages, aligning them with the information specified in the accompanying JSON file.
Upon completion of the printing process, a Universal Robot UR10 robotic arm transfers the surface, as shown in Fig. \ref{fig:tranfer_printer_drone}, along with the printed components, onto a drone positioned on a designated landing pad adjacent to the printer. The magnetic surface attaches securely to the drone's gripping mechanism, ensuring stable retention of the components during flight and safe transport to the next phase of the workflow. In Fig. \ref{fig:teaser}(b), the part production station is shown. 

\subsection{Flying conveyor stage} 

A swarm of drones within the concept of Industry 6.0 can be utilized to transport objects between key points in the system. Generative AI analyzes the current state of the production line and makes decisions on the implementation of appropriate tasks. The presented prototype consists of an 8-inch custom drone equipped with a SpeedyBee F405 flight controller, which is based on ArduPilot autopilot firmware with a MAVROS interface (shown in Fig. \ref{fig:teaser}(c)). It is equipped with an onboard OrangePi 5B computer, which is used for sending control data to the flight controller and processing the current state of the production line. A Vicon Tracking system with 14 infrared cameras is used for precise localization, providing high-quality tracking of the drone's position and movement. Drones are sent to the target flight path and positions with PID control parameters. The Robot Operating System (ROS) framework is applied to run the developed software packages. The system includes a PC running the Vicon framework and a drone-control framework integrated with a decision-making system to facilitate autonomous operations. The drone was designed to demonstrate a system for transporting objects within a production line. It is equipped with a mechanism designed to transport a 3D printer surface that holds various parts. The printing surface is attached to the drone using magnets, enabling stable and efficient transport. The locations where the drone needs to land precisely are equipped with special landing platforms to ensure accurate positioning.

\subsection{Product assembling stage} 



\begin{figure}[t]
\vspace{+0.4cm}
\centerline{\includegraphics[width=0.48\textwidth]{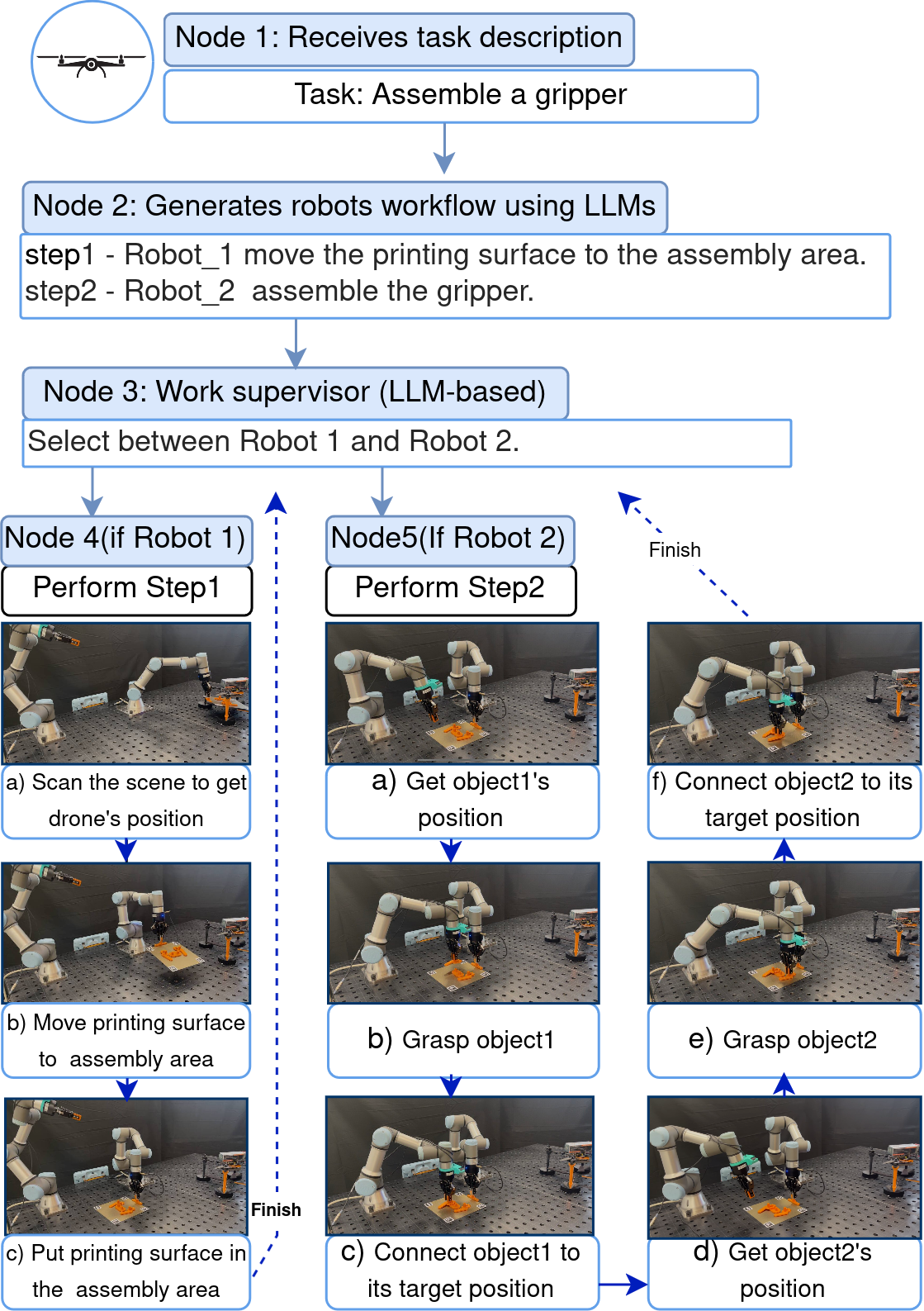}}
\caption{LangGraph-inspired assembly pipeline of the parts delivered by drones.}
\label{fig:assemly_area}
\vspace{-0.4cm}
\end{figure}

At this stage, two robot manipulators, UR3 from Universal Robots, are tasked with detaching the printed parts from the printing surface and assembling the final components. Each robot is equipped with a 2-finger Robotiq gripper 2F-83 and operates sequentially. 
The product assembling area is shown in Fig. \ref{fig:teaser}(d).
The orchestration of the workflow between the two robots is managed by the LLMs through a structured graph using LangGraph. Following the general workflow in the form of steps from the LLM, the first robot is responsible for taking the components from the drone, bringing them to the assembly area, and, if needed, returning the final product to the drone. The second robot is tasked with detaching the parts from the printing surface and assembling the components. Fig. \ref{fig:assemly_area} shows the detailed graph structure. 
An LLM-based supervisor then assigns each step of the workflow to the appropriate robot based on its knowledge of the ability of each robot provided in the prompt.
After the selection of the appropriate robot, a new node in the graph is responsible for the generation of a list of subtasks (see a)-c) for robot 1 and a)-f) for robot 2) to accomplish the task. 
Each subtask triggers the selection of appropriate predefined API functions (for controlling the motions of the robots and opening and closing of the gripper) with appropriate variables like object names as well as the objects' positions, which are primarily provided in a JSON file generated in the first stage (see subsection \ref{blueprint_generation_stage}).

\section{Experiments} 

\subsection{LLM Generation comparison}

To evaluate the capabilities of various LLMs in generating mechanism assemblies, we conducted an experiment using our LangChain-based pipeline. The assessment focused on three key aspects: mechanism analysis from user input, generation of executable Python code for SDF mechanism assembly, and functional mechanism STL file production.

\textbf{Procedure}: the experiment involved ten different mechanisms of varying complexity, ranging from simple ‘text with a stick’ to more intricate designs like ‘pliers’. For each mechanism, we provided a description tailored to its complexity, with simpler mechanisms receiving concise inputs and more complex ones accompanied by detailed descriptions of primitives, part interpositions, and general assembly information.

We compared nine LLMs with diverse characteristics, including parameter count, context window size, and coding and reasoning abilities. Among these were four top-tier commercial models: the state-of-the-art OpenAI o1-preview model \cite{o1preview}, which employs reinforcement learning for complex reasoning tasks, GPT-4o \cite{gpt4o}, Claude 3.5 Sonnet by Anthropic \cite{sonnet}, and Google Gemini 1.5 Pro \cite{team2024gemini}. In addition to these top-performing models, we evaluated smaller and faster commercial LLMs like GPT-4o-mini \cite{gpt4omini} and Claude 3 Haiku \cite{haiku}. We also tested the open-source LLMs based on Meta LLama 3.1 \cite{dubey2024llama}, including Hermes 3 \cite{teknium2024hermes} 405B — neutrally-aligned generalist instruct and tool use model with strong reasoning abilities. Finally, we tested our pipeline locally, leveraging Gemma 2 9B \cite{Riviere2024Gemma2I} by Google DeepMind. Each model received the user input and was tasked with producing a valid 2D SDF mechanism assembly.

The evaluation employed a binary scoring system across three categories. In the Analysis, the model's ability to produce a detailed description of the mechanism assembly, the decomposition of the mechanism into its parts, and the connection between the parts were evaluated. In the Executable Code category, the LLM's capacity to generate Python code for the 2D SDF mechanism assembly, adhering to the rules and examples provided in the LLM context was evaluated. In the Mechanism STL, the final assembly STL was assessed. Since the final output entirely relies on 2D SDF assembly, generated in the previous sub-stage, the LLM ability to produce a code for the functional machine is of most importance.

\textbf{Results}: the experimental results are depicted in Table \ref{tab:llm-comparison}.

\begin{table}[htbp]
\caption{Comparison of LLMs across Generation Steps}
\label{tab:llm-comparison}
\centering
\resizebox{\columnwidth}{!}{%
\begin{tabular}{|l|c|c|c|}
\hline
\textbf{LLMs} & \multicolumn{3}{c|}{\textbf{Generation Steps}} \\
\hline
 & \textbf{Analysis} & \textbf{Executable Code} & \textbf{Mechanism STL} \\
\hline
\textbf{o1-preview} & \textbf{10} & \textbf{10} & \textbf{8} \\
\textbf{gpt-4o} & \textbf{10} & \textbf{10} & \textbf{8} \\
claude-3.5-sonnet & 10 & 9 & 8 \\
gemini-1.5-pro-exp & 9 & 8 & 5 \\
hermes-3-llama-3.1-405b & 9 & 6 & 5 \\
gpt-4o-mini & 9 & 9 & 4 \\
reflection-llama-3.1-70b & 7 & 6 & 2 \\
claude-3-haiku & 7 & 4 & 2 \\
gemma-2-9b & 7 & 4 & 1 \\
\hline
\end{tabular}%
}
\end{table}

\begin{figure*}[htbp]
\vspace{+0.3cm}
  \centering
\includegraphics[width=0.95\textwidth]{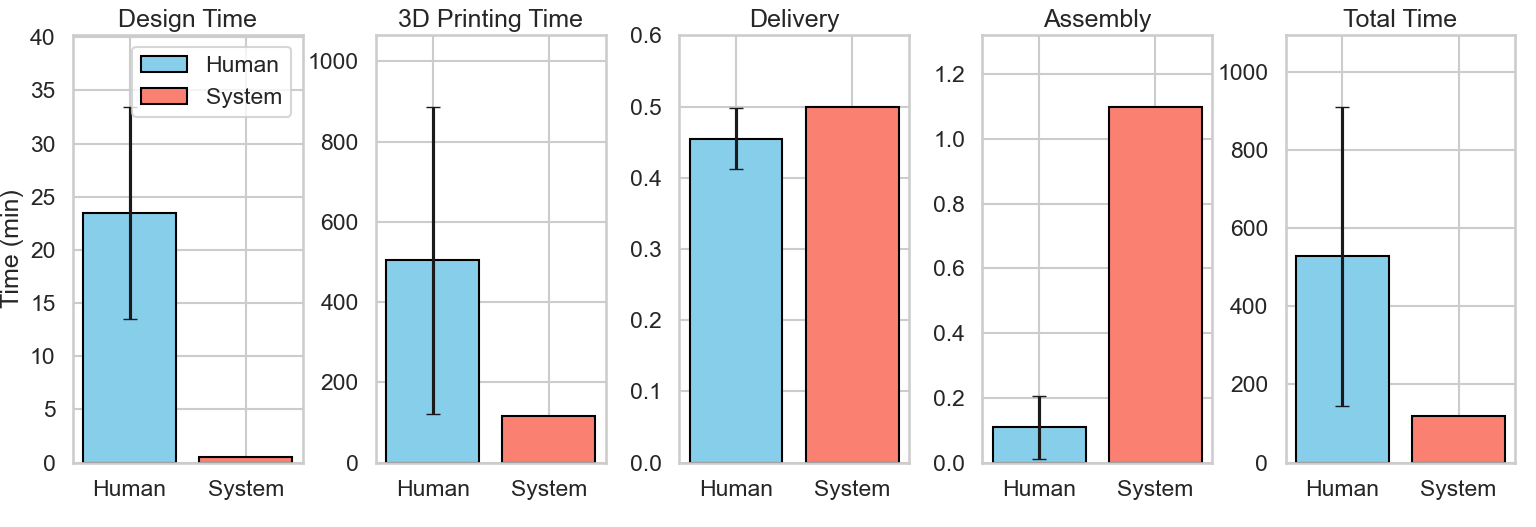}
\caption{Comparison of the time spent on the development and production of a test part by human subjects and the proposed automated system under the supervision of the LLM.}
\label{fig:time_comparision}
\vspace{-0.3cm}
\end{figure*}

According to the evaluation, OpenAI's o1-preview and gpt-4o models demonstrated superior performance across all three categories, achieving perfect scores in Analysis and Executable Code generation, and the highest score in Mechanism STL production. Claude-3.5-sonnet showed comparable performance to the top models, with only a slight lag in Executable Code generation. Smaller models like claude-3-haiku and gemma-2-9b struggled with the more complex aspects of the task, particularly in Executable Code and Mechanism STL generation.

However, all models exhibited a noticeable decline in performance as task complexity increased, with the Mechanism STL category emerging as the most challenging.

This experiment highlights the LLMs' capabilities for the complex task of mechanism assembly generation. According to experimental results, leading models like GPT-4o should be used for a reliable outcome.

\subsection{User study}

To assess the efficiency of our system, we conducted a user study with participants tasked with manufacturing a gripper mechanism through four stages. 
Each user started with the design of the gripper, followed by 3D printing of the designed model, delivering the printed parts to the assembly area, and concluding the assembly. We recorded the total time participants took to complete the process and the completion time at each stage and compared it with the system's autonomous performance.

\textbf{Subjects}: ten participants with a background in mechanical design and 3D printing, aged 23 to 34 years (26 $\pm$3.9 years), participated in the experiment.

\textbf{Procedure}: before the experiment, the task was thoroughly explained to each participant. Subjects were asked to design a gripper mechanism consisting of three parts: a base and two fingers, with pins and holes for assembly, and were given the freedom to utilize any preferable CAD tools. After designing, participants were asked to export the assembled mechanism as an STL file and proceed with 3D printing of all parts. Once the printing stage was complete, they were to deliver the parts to the same assembly table used by the automatic system and assemble the mechanism. The design, printing, and assembly areas were all situated within a 6-square-meter space. The time required by each participant to complete each stage was recorded for subsequent analysis.

\textbf{Results}: the results of the user study are illustrated in Fig. \ref{fig:time_comparision}. The results demonstrated that the system outperformed the human participant in manufacturing speed by a factor of 4.44 (528.64 min for the human versus 119.10 min for the system). This acceleration was primarily due to substantial reductions in time during the design stage (from 23.50 min to 0.50 min, a factor of 47) and the 3D printing stage (from 504.57 min to 117.00 min, a factor of 4.3) enabled by LLM-driven shape generation optimized for printing. In contrast, component delivery times were nearly identical (0.46 min for the human vs. 0.50 min for the system, with a marginal 8.7\% advantage for the human). However, the human participant completed assembly more efficiently (0.11 min vs. 1.10 min for the system, a factor of 10 faster).

We anticipate that as part complexity increases and the distance between manufacturing and assembly areas grows, the advantage in delivery and assembly times may shift in favor of an LLM-supervised automated approach.

\section{Conclusion and Future Work}

This paper introduces the concept of Industry 6.0, presenting the world’s first fully autonomous production system capable of managing the entire design and manufacturing process based on natural language descriptions provided by users. A swarm of heterogeneous robots orchestrates the production process in four stages: blueprint creation, part manufacturing, logistics, and assembly, each stage leveraging generative AI models. The proposed system was tested and evaluated using both commercial and open-source LLMs, including API-based services and local inference.

The system demonstrated remarkable performance in producing multi-component products following user commands. State-of-the-art models in the analysis stage achieved a 10/10 success rate in generating accurate product descriptions. At the final stage, 8/10 product descriptions were successfully converted into valid STL models, which were then autonomously produced by the system. A comprehensive user study further highlighted the system’s efficiency, reducing the average production time to 119.10 minutes, significantly outperforming a team of expert human developers who averaged 528.64 minutes, an improvement by a factor of 4.4. Moreover, in the blueprinting stage, the system outpaced human CAD operators by an unprecedented factor of 47, completing the task in just 0.5 minutes compared to the human average of 23.5 minutes. This breakthrough marks a significant step toward fully autonomous manufacturing.

Looking ahead, the industrial potential of this technology can evolve in two primary directions. The first path focuses on creating autonomous production systems capable of manufacturing personalized products on demand. The second path envisions full autonomy, where the system autonomously analyzes market trends and optimizes production processes to manufacture the most in-demand or profitable products based on strategic objectives. Future work will explore these possibilities, pushing the boundaries of autonomous production systems.

\bibliographystyle{IEEEtran}
\bibliography{mybib}

\end{document}